# Optimization of MedSAM model based on bounding box adaptive perturbation algorithm

Boyi Li[1][0009-0008-1370-7474], Ye Yuan[2], Wenjun Tan[3]

[1,2,3]School of Computer Science and Engineering, Northeastern University

*Abstract*—The MedSAM model, built upon the SAM framework, enhances medical image segmentation through generalizable training but still exhibits notable limitations. First, constraints in the perturbation window settings during training can cause MedSAM to incorrectly segment small tissues or organs together with adjacent structures, leading to segmentation errors. Second, when dealing with medical image targets characterized by irregular shapes and complex structures, segmentation often relies on narrowing the bounding box to refine segmentation intent. However, MedSAM's performance under reduced bounding box prompts remains suboptimal. To address these challenges, this study proposes a bounding box adaptive perturbation algorithm to optimize the training process. The proposed approach aims to reduce segmentation errors for small targets and enhance the model's accuracy when processing reduced bounding box prompts, ultimately improving the robustness and reliability of the MedSAM model for complex medical imaging tasks.

*Keywords*—MedSAM, Medical Image Segmentation, Bounding Box Perturbation, Generalizable Training, Segmentation Accuracy

## I. INTRODUCTION

Accurate identification and segmentation of tissues and structures in medical images are essential for disease diagnosis, treatment planning, and monitoring therapeutic outcomes. Recent advancements have introduced MedSAM [1], a generalization model built upon the SAM architecture, specifically tailored for medical image segmentation. MedSAM not only preserves the strengths of SAM in zero-shot generalization but also significantly enhances segmentation performance across medical imaging tasks. Empirical evidence demonstrates that MedSAM outperforms existing state-of-the-art (SOTA) segmentation models [2] and achieves comparable, if not superior, results to specialized models [3],[4] designed for the same imaging modalities.

Despite these advancements, MedSAM relies on high-quality, standardized boundary prompt information to ensure accurate segmentation results in medical image processing. When this prompt information is compromised or biased, the model's segmentation performance can degrade significantly. To address this issue, existing research has pursued two primary strategies: the first involves enhancing segmentation accuracy by incorporating multiple sources of prompt information, while the second focuses on improving the model's robustness by adopting advanced network architectures or composite models designed to accommodate deviations in the provided prompt data. These approaches aim to mitigate the impact of unreliable or incomplete cues and ensure more reliable segmentation outcomes.

In the domain of prompt enhancement, several studies have explored innovative strategies. Deng et al. [5] proposed a multi-frame prompting scheme to evaluate and reduce prompting uncertainty. Li et al. [6] developed a 3D medical image segmentation model that integrates single-point prompts with a SAM pre-trained visual converter, leading to notable improvements in segmentation accuracy. Wu and Xu [7] introduced a method combining one-time prompting and interactive prompting to adapt flexibly to diverse segmentation tasks. Zhou et al. [8] constructed an uncertainty-driven prompting framework for MedSAM, incorporating multiple uncertain prompts to enhance model performance. Xu et al. [9] designed a self-patch prompt generator capable of automatically producing high-precision dense prompt embeddings, achieving efficient self-prompting for generalized medical imaging. Liu et al. [10] proposed an iterative framework to transform point prompts into pseudo bounding box suggestions, improving MedSAM's segmentation capability under semantically-aware point supervision. Gaillochet, Desrosiers and Lombaert [11] developed a prompt embedding learning module based on image embedding, enabling MedSAM to automate region-specific segmentation tasks. Khaertdinova et al. [12] employed a semi-supervised approach utilizing human gaze data as prompts to enhance segmentation under uncertain prompting conditions. Zhao and Shen [17] introduced a part-aware prompting mechanism, leveraging part-level features and multi-point prompts to reduce uncertainty and improve segmentation accuracy. Marinov et al. [19] optimized MedSAM for thin structures, such as blood vessels, using doodle-based pixel-level fine prompts.

Parallel efforts in model architecture optimization have yielded promising results. Hu et al. [13] proposed a context learning framework that combines SAM and ICL models to generate prompts via ICL, eliminating the reliance on manual prompts. Deshpande et al. [14] utilized weakly labeled models trained on standard labels to generate automated prompts for unlabeled medical images, simplifying segmentation and improving usability. Fu et al. [15] designed a self-correcting SAM framework that generates rough segmentation masks in unprompted mode, eliminating the need for a prompt generator while providing preparatory prompts for optimization. Ayzenberg, Giryes and Greenspan [16] combined the ALPnet prototype network with the DINOv2 encoder to refine coarse masks, ensuring consistent and accurate segmentation results. Magg, Kervadec and Sanchez [18] proposed a non-iterative optimal prompting strategy using bounding boxes, points, and combinatorial prompts to validate SAM's zero-shot capability in skeletal CT segmentation.

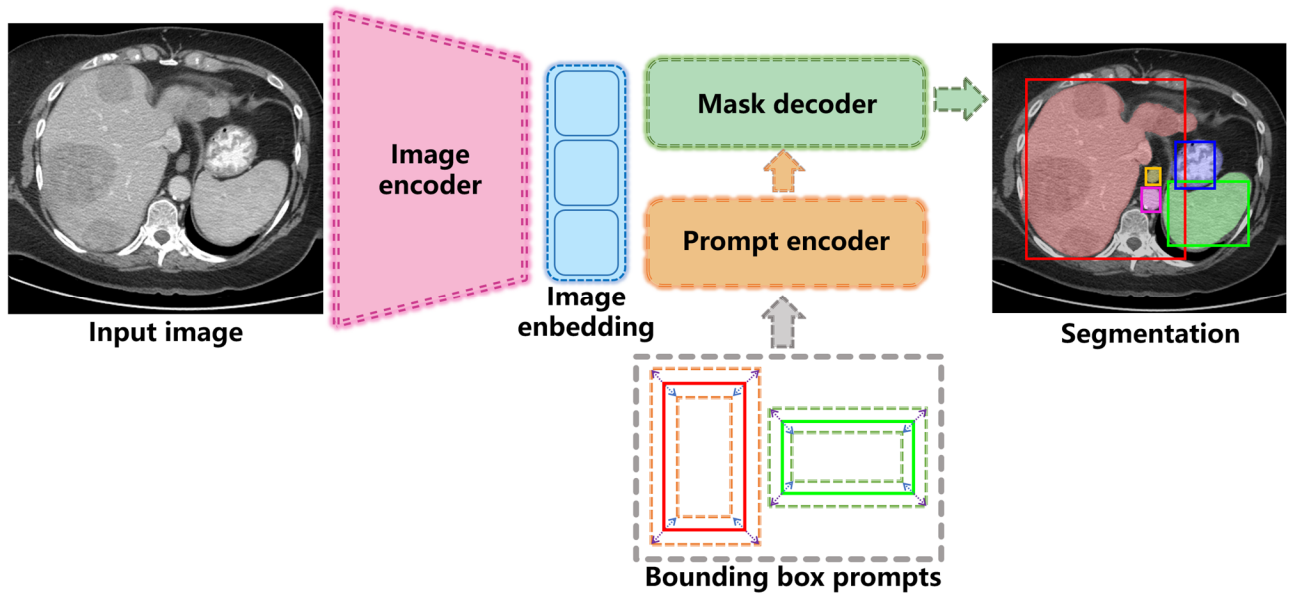

Fig. 1. Optimized MedSAM network architecture

Although the aforementioned studies have enhanced the reliability and accuracy of MedSAM in practical applications, the reliance on extensive boundary hint annotations to reduce uncertainty presents significant challenges. This approach not only increases the workload for clinicians during the annotation process but also results in an excessive amount of boundary hint information that the model must process, severely affecting its operational efficiency. Moreover, the inherent complexity of the MedSAM model further complicates its application. The use of composite model self-prompting methods to define segmentation domains risks exacerbating this complexity, potentially introducing redundancy and reducing the model's efficiency in real-world settings.

To address the challenges identified in current research, this study proposes improvements to the model's prompt perturbation settings through the introduction of a bounding box adaptive perturbation algorithm. As illustrated in Fig. 1, the algorithm incorporates a novel boundary prompt perturbation strategy during the training process. This approach is designed to account for potential biases in the boundary prompts provided by clinicians in practical applications, thereby enhancing the model's ability to effectively process boundary prompt information with perturbations.

Specifically, this study makes the following contributions:

(1) **Quantification of Feature Relationships:** We introduce a scale coefficient for the segmentation domain and a similarity coefficient between the segmentation domain and the global domain. These coefficients are used to quantify the relationship between the region of interest in the binarized sample data and the broader global domain across multiple dimensions. The algorithm applies random perturbations in different dimensions to resolve the challenge of setting an appropriate threshold for perturbation, thereby improving the accuracy of the segmentation process.

(2) **Adaptive Perturbation Factors:** The algorithm introduces both a perturbation expansion factor and a perturbation shrinkage factor to adapt to the variations in bounding box prompt information. These factors allow the model to better handle random perturbations associated with shrinking bounding boxes, thereby improving segmentation accuracy. Additionally, this adaptation enhances the model's generalization ability, making it more robust to diverse perturbation scenarios in practical applications.

II. METHOD

In this study, we propose improvements to the prompt engineering design of the MedSAM model, while retaining its original network architecture. No modifications are made to the core structure of MedSAM. The code for the original model is publicly available at https://github.com/bowang-lab/MedSAM, ensuring accessibility and reproducibility. Our focus is on enhancing the prompt engineering process to improve the model's performance in medical image segmentation tasks.

A. Bounding box adaptive perturbation algorithm

This study proposes an adaptive bounding box perturbation optimization algorithm designed to dynamically adjust the perturbation threshold. The goal is to balance the sensory field and enhance the model's segmentation accuracy and generalization capability. As illustrated in Fig. 2, the algorithm introduces a global similarity coefficient, $\theta_\omega$, between the segmentation domain and the image. This coefficient allows the boundary prompt to adaptively adjust the perturbation range based on the relative size of the region of interest within the overall image. This mechanism helps prevent excessive perturbation when segmenting small targets. Additionally, a boundary scale factor, $\xi$, is introduced to preserve the geometric similarity of the frame-embedded prompt information before and after perturbation. This factor is crucial for accurately segmenting small tissues, where maintaining geometric consistency is essential.

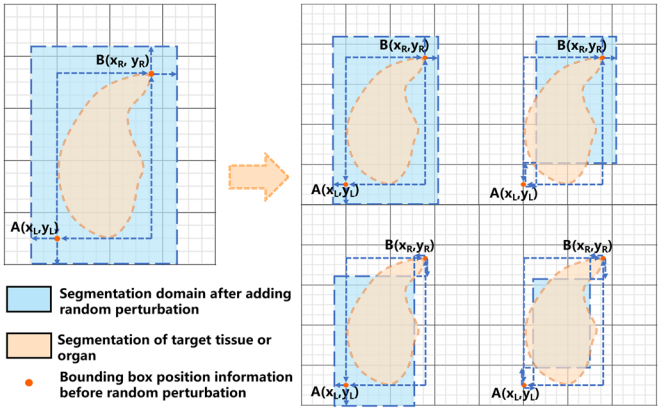

Fig. 2. Random perturbation segmentation domain comparison before and after optimization

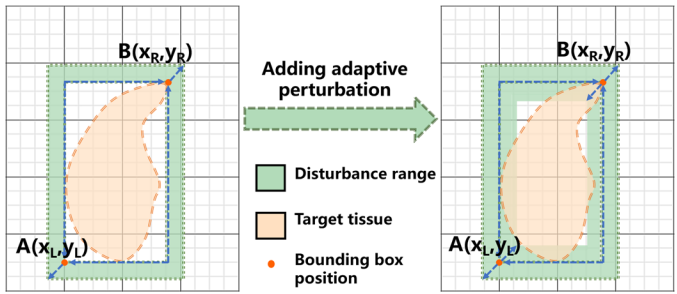

Fig. 3. Comparison of boundary box perturbation range before and after optimization

In the design of the boundary perturbation trend, the proposed optimization algorithm enhances the model's responsiveness to boundary prompt uncertainty by introducing two perturbation factors: the expanding perturbation factor $\delta_{expand}$ and the shrinking perturbation factor $\epsilon_{shrink}$. These factors enable bidirectional perturbation of the bounding box during model training, allowing the expanding perturbation to occur simultaneously with the shrinking perturbation, as shown in Fig. 3. To prevent overfitting caused by excessive model sensitivity, the algorithm sets maximum and minimum limits for the perturbations: the maximum expanding perturbation $\delta_{expand}^{max}$ and the minimum shrinking perturbation $\epsilon_{shrink}^{min}$. These constraints ensure that the model's sensitivity to boundary perturbations remains within a defined range.

The 2D perturbation offsets $(\epsilon_1, \epsilon_2, \delta_1, \delta_2)$ formula for this algorithm are calculated using the following formulas:

$$\epsilon_1 = \epsilon_{shrink} \cdot \theta_\omega \ (\epsilon_{shrink} \geq \epsilon_{shrink}^{min}) \quad (1)$$

$$\epsilon_2 = \epsilon_{shrink} \cdot \theta_\omega / \xi \ (\epsilon_{shrink} \geq \epsilon_{shrink}^{min}) \quad (2)$$

$$\delta_1 = \delta_{expand} \cdot \theta_\omega \ (\delta_{expand} \leq \delta_{expand}^{max}) \quad (3)$$

$$\delta_2 = \delta_{expand} \cdot \theta_\omega / \xi \ (\delta_{expand} \leq \delta_{expand}^{max}) \quad (4)$$

Using the standard bounding box parameters $(\alpha_1, \alpha_2, \beta_1, \beta_2)$, the perturbed bounding box coordinates $(\alpha_1', \alpha_2', \beta_1', \beta_2')$ are generated through the adaptive perturbation optimization as follows:

$$\alpha_1' \sim U(\alpha_1 + \epsilon_1, \alpha_1 + \delta_1) \quad (5)$$

$$\alpha_2' \sim U(\alpha_2 + \epsilon_2, \alpha_2 + \delta_2) \quad (6)$$

$$\beta_1' \sim U(\beta_1 + \epsilon_1, \beta_1 + \delta_1) \quad (7)$$

$$\beta_2' \sim U(\beta_2 + \epsilon_2, \beta_2 + \delta_2) \quad (8)$$

*B. loss function*

In this study, we use the unweighted sum between the cross-entropy loss and the dice loss as the final loss function, as this combination has demonstrated robustness across various medical image segmentation tasks. Specifically, let $S$ and $G$ represent the predicted segmentation and the ground truth, respectively. $s_i$ and $g_i$ denote the predicted segmentation and ground truth value for voxel $i$, and $N$ is the total number of voxels in the image $I$. The binary cross-entropy loss is defined as:

$$L_{BCE} = -\frac{1}{N} \sum_{i=1}^{N} [g_i \log s_i + (1-g_i) \log(1-s_i)] \quad (9)$$

Dice loss is defined as:

$$L_{Dice} = 1 - \frac{2 \sum_{i=1}^{N} g_i s_i}{\sum_{i=1}^{N} (g_i)^2 + \sum_{i=1}^{N} (s_i)^2} \quad (10)$$

The total loss function $L$ is the sum of the binary cross-entropy and Dice loss:

$$L = L_{BCE} + L_{Dice} \quad (11)$$

To prevent model overfitting, we incorporate weight decay (WD) as a regularization technique. This method introduces an additional penalty term to the loss function, which limits the size of the model parameters, thereby reducing overfitting to the noise present in the training data. The final loss function is then defined as:

$$L_{final} = L + \lambda \cdot \frac{1}{2} \sum_i w_i^2 \quad (12)$$

III. EXPERIMENT

*A. Dataset*

The datasets used in this study include the MICCAI FLARE22 challenge dataset, which is focused on abdominal segmentation across 13 organ classes, and a medical imaging dataset of lung tissues from in-house dataset. The MICCAI FLARE22 dataset consists of 50 abdominal CT scans, each annotated with segmentation masks for 13 different organs. The in-house dataset includes 24 CT image sets of lung tissue.

For data preprocessing, the window width and position for the FLARE22 CT images were adjusted to a range of -360 to 440, while for the lung CT images, the range was set to -1000 to 400. Both datasets were then resampled to a resolution of 1024*1024 pixels and normalized using min-max normalization. The datasets were randomly split into training, validation, and testing sets, with 80% of the data used for training, 10% for validation, and 10% for testing.

*B. Experiment settings*

The experiments were conducted using Python 3.12.1, Torch 2.2.1, and an NVIDIA 4060 laptop with a 16 GB GPU. The model was trained for 20 epochs using the AdamW optimizer, with an initial learning rate of 0.0001. To enhance the training process, the ReduceLROnPlateau learning rate scheduler was employed to dynamically adjust the learning rate based on performance. Due to hardware limitations, the batch size was set to 1, with each batch consisting of one labeled and one unlabeled image. The patch size was set to 1024*1024*3.

To optimize training efficiency and reduce the time required for each epoch, we utilized automatic mixed-precision (AMP) during GPU training. AMP combines both 16-bit and 32-bit floating-point arithmetic to enhance computational performance while minimizing memory usage, thus accelerating training and improving overall efficiency.

*C. Evaluation Metrics*

This study uses Dice Similarity Coefficient (DSC) and Normalized Surface Distance (NSD) to quantitatively evaluate segmentation results. DSC is a region-based segmentation metric designed to evaluate the region overlap between the expert annotation mask and the segmentation result, which is defined as:

$$DSC(G,S) = \frac{2|G \cap S|}{|G| + |S|} \quad (13)$$

NSD is a boundary-based metric designed to evaluate the boundary consistency between the expert annotation mask and the segmentation result for a given tolerance, which is defined as:

$$NSD(G,S) = \frac{|\delta G \cap B_{\delta S}^{(\tau)}| + |\delta S \cap B_{\delta G}^{(\tau)}|}{|\delta G| + |\delta S|} \quad (14)$$

*D. Results*

This study compares the segmentation performance of three models: SAM [20], LiteMedSAM [21], and Optimized MedSAM, each processing segmentation tasks with boundary prompts containing different levels of perturbation. As shown in Table 1, Optimized MedSAM achieves the highest mean scores, with a DSC of 93.87% and an NSD of 95.50%. In contrast, SAM achieves DSC and NSD scores of 90.33% and 91.82%, respectively, while LiteMedSAM obtains DSC and NSD scores of 88.11% and 90.01%, respectively. The segmentation results are illustrated in Fig. 4.

The Optimized MedSAM model incorporates perturbation prompts during training to simulate the uncertainties typically encountered in real-world scenarios. These perturbations enable the model to learn deeper feature associations between the prompt and the target region, improving its ability to handle uncertainty. Through joint learning with the ground truth labels, the model gradually adapts to these perturbations, enhancing its robustness to boundary variations and significantly expanding the effective range of boundary prompts.

TABLE I. THE SEGMENTATION EFFECT OF DIFFERENT MODELS ON TASKS WITH DISTURBANCE PROMPT

| Boundary shift | SAM | | Lite MedSAM | | Optimized MedSAM | |
|---|---|---|---|---|---|---|
| | DSC(%) | NSD(%) | DSC(%) | NSD(%) | DSC(%) | NSD(%) |
| Standard | 93.92 | 95.12 | 94.06 | 95.76 | **95.67** | **96.78** |
| Expansion | 91.66 | 93.04 | 92.11 | 93.89 | **94.71** | **96.13** |
| Shrink | 85.42 | 87.29 | 78.17 | 80.39 | **91.24** | **93.58** |
| Average | 90.33 | 91.82 | 88.11 | 90.01 | **93.87** | **95.50** |

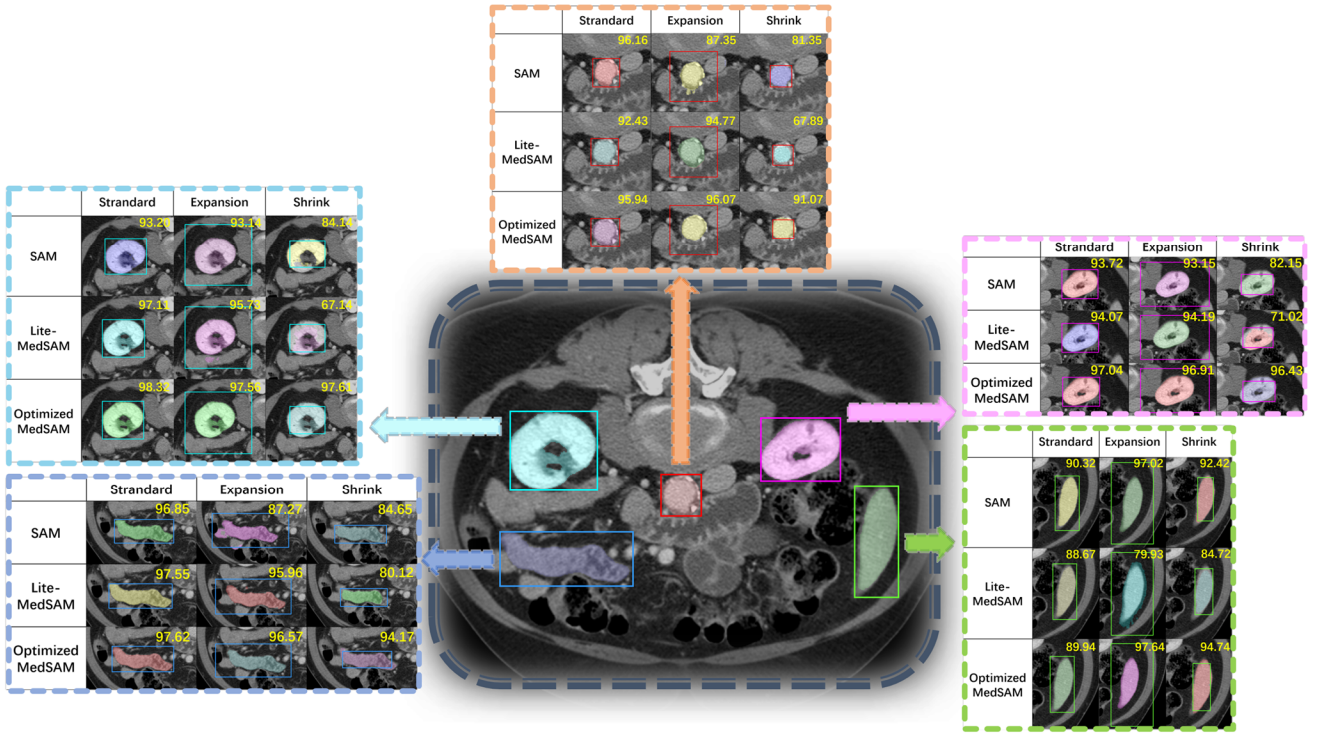

Fig. 4. The segmentation effect of different models in different disturbance prompts

*E. Ablation studies*

This study conducts ablation experiments to evaluate the impact of the proposed optimization algorithm by comparing the model's performance before and after optimization. The following two tasks are tested:

(1) Segmentation of densely distributed microtissues to assess the algorithm's effectiveness in improving segmentation of small, closely-packed structures.

(2) Segmentation with boundary prompts that involve narrowing perturbation regions, aimed at evaluating the algorithm's ability to enhance the model's adaptation to perturbation variations.

**Optimization effectiveness of the bounding box adaptive perturbation algorithm for segmenting tiny tissues:**

As shown in Fig. 5, when the pre-optimization model receives box-embedded prompts with a smaller perturbation range, it often includes non-target tissues from the surrounding area in the segmentation result. This issue is particularly pronounced when the target region's boundary is ambiguous or neighboring tissues are densely distributed.

This behavior arises because the random perturbation values fixed during the training of the MedSAM model do not adequately account for the characteristics of small-scale targets. Specifically, during training, the model maps the box-embedded prompts, which may be far from the actual target, to the true segmentation region. For small targets, the random perturbation's offset can be too large, leading the model to over-respond to the perturbation. As a result, the model perceives an exaggerated expansion of the bounding box, causing segmentation errors. This highlights the necessity of a more adaptive perturbation mechanism, especially for tiny tissues.

The optimization algorithm proposed in this study modifies the perturbation strategy by introducing two sets of scheduling factors. This adjustment effectively limits the over-expansion of the model's sensory field, preventing neighboring tissues outside the bounding box from being incorrectly included in the segmentation. As shown in Table 2, the error rate for segmenting tiny tissues decreases from 18% to 2% with the optimized model, significantly enhancing its robustness and accuracy in small-scale target segmentation.

**Optimization effectiveness of the bounding box adaptive perturbation algorithm for reduced perturbation:**

This study evaluates the performance of the pre-optimization and post-optimization models on segmentation tasks with shrinking perturbation prompt information. As shown in Fig. 6, the pre-optimization model exhibits significant deviations from the true label values when handling shrinking perturbation prompts. In contrast, the post-optimization model generates segmentation results that are closer to the true labels under the same conditions.

As presented in Table 3, after incorporating the two sets of perturbation scheduling factors, the model achieves a significant improvement in performance. The average Dice Similarity Coefficient (DSC) increases from 73.42% to 90.72%, and the Normalized Surface Distance (NSD) improves from 75.27% to 92.86% for the task involving reduced perturbation prompts. These results demonstrate that the optimized model is more adaptive and robust to prompt information under conditions of reduced perturbation.

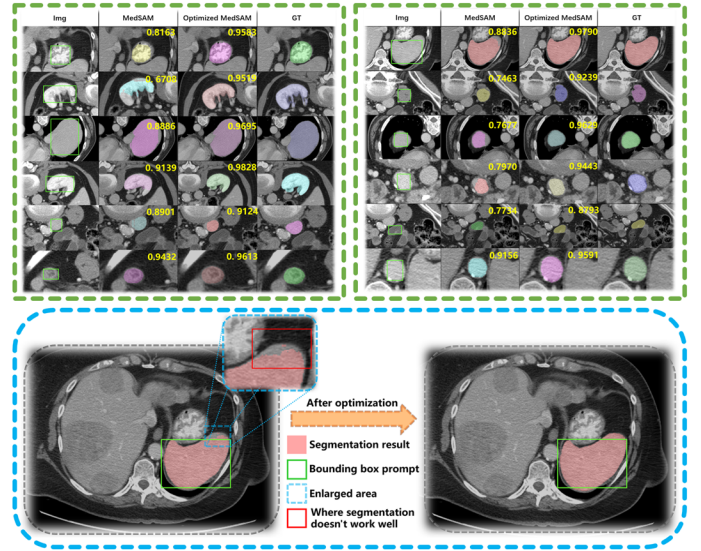

Fig. 6. Optimization effect when dealing with reduced disturbance

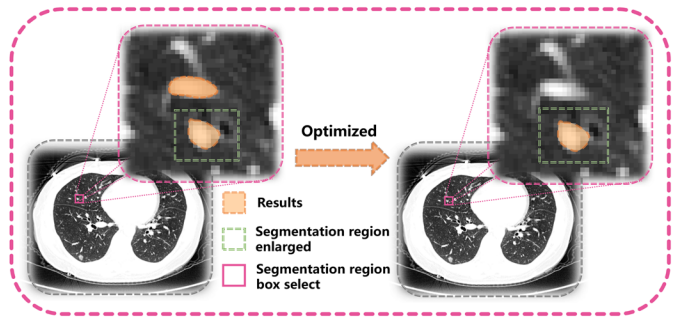

Fig. 5. Micro-tissue segmentation effect before and after model optimization

TABLE II. THE ERROR RATE OF SEGMENTATION OF SMALL TISSUE

| MedSAM | Add $\theta_\omega, \xi$ | Add $\delta_{expand}, \epsilon_{shrink}$ | Error Rate |
|---|---|---|---|
| ✓ | | | 18% |
| ✓ | | ✓ | 16% |
| ✓ | ✓ | | 6% |
| ✓ | ✓ | ✓ | **2%** |

TABLE III. SEGMENTATION ACCURACY OF THE MODEL BEFORE AND AFTER OPTIMIZATION FOR SEGMENTATION JOBS WITH SHRINK DISTURBANCE

| MedSAM | Add $\theta_\omega, \xi$ | Add $\delta_{expand}, \epsilon_{shrink}$ | DSC | NSD |
|---|---|---|---|---|
| ✓ | | | 73.42 | 75.27 |
| ✓ | ✓ | | 74.35 | 75.24 |
| ✓ | | ✓ | 87.42 | 89.73 |
| ✓ | ✓ | ✓ | **90.72** | **92.86** |

*F. Limitation*

The perturbation prompt information provided by the operator is inherently subjective, which results in the absence of a clear threshold for distinguishing between reasonable and erroneous perturbations. Variations in the criteria for defining the reasonableness of perturbations can directly impact the evaluation of the algorithm's optimization effectiveness. Consequently, quantifying the precise effectiveness of the algorithm in optimizing existing models remains a challenge. However, it is important to note that, within a certain range, increasing the threshold for defining perturbation reasonableness generally enhances the optimization effect of the algorithm.

## IV. CONCLUSION

This study introduces a bounding box adaptive perturbation optimization algorithm based on the MedSAM model. The algorithm dynamically adjusts the perturbation range of the bounding box by considering the geometric features of the regions of interest (ROIs) and their relative size to the image. Specifically, it addresses two key aspects: first, it prevents the model from excessively expanding its sensory field for small boundary prompts, thereby improving segmentation accuracy for tiny tissues or organs; second, it maintains high segmentation performance under uncertain, shrinking boundary prompts by introducing a controlled, bi-directional perturbation threshold. These improvements enhance the robustness and accuracy of the MedSAM model in challenging segmentation tasks.

The findings have important implications for clinical medical image segmentation. The algorithm not only improves accuracy for tiny tissue segmentation, providing clinicians with more reliable decision-making tools and enhancing surgical planning and execution, but also reduces annotation and adjustment workloads in clinical studies by incorporating a realistic perturbation mechanism during training. Furthermore, it enhances the model's adaptability to boundary prompts with perturbations, ensuring accurate, robust, and stable segmentation across diverse clinical scenarios.